\documentclass[10pt,twocolumn,letterpaper]{article}

\usepackage{cvpr}              

\usepackage{graphicx}
\usepackage{amsmath}
\usepackage{amssymb}
\usepackage{booktabs}

\usepackage[pagebackref,breaklinks,colorlinks]{hyperref}

\usepackage[capitalize]{cleveref}
\crefname{section}{Sec.}{Secs.}
\Crefname{section}{Section}{Sections}
\Crefname{table}{Table}{Tables}
\crefname{table}{Tab.}{Tabs.}

\def\cvprPaperID{9} 
\def\confName{CVPR}
\def\confYear{2025}

\begin{document}

\title{Accurate Automatic 3D Annotation of Traffic Lights and Signs for Autonomous Driving}

\author{Sándor Kunsági-Máté \and Levente Pető \and Lehel Seres \and Tamás Matuszka\\
{\href{https://aimotive.com}{aiMotive}}\\
{\tt\small \{sandor.kunsagimate, levente.peto, lehel.seres, tamas.matuszka\}@aimotive.com}
}
\maketitle

\begin{abstract}
3D detection of traffic management objects, such as traffic lights and road signs, is vital for self-driving cars, particularly for address-to-address navigation where vehicles encounter numerous intersections with these static objects. This paper introduces a novel method for automatically generating accurate and temporally consistent 3D bounding box annotations for traffic lights and signs, effective up to a range of 200 meters. These annotations are suitable for training real-time models used in self-driving cars, which need a large amount of training data. The proposed method relies only on RGB images with 2D bounding boxes of traffic management objects, which can be automatically obtained using an off-the-shelf image-space detector neural network, along with GNSS/INS data, eliminating the need for LiDAR point cloud data.
\end{abstract}

\section{Introduction}
\label{sec:intro}

Autonomous driving is currently one of the most actively researched fields. Given the complexity of the problem, recent advancements focus on perceiving the entire three-dimensional environment around the vehicle. This comprehensive approach is essential because of the myriad traffic scenarios and interdependencies between objects, making two-dimensional object detection insufficient due to the lack of depth information. For instance, detecting a red light in a self-driving car's camera image does not necessarily mean the vehicle must stop. How far away is the traffic light? Is it relevant to the lane in which the ego vehicle is located? To answer these questions, the 3D positions of the objects have to be known.

Deep learning models currently used in self-driving cars require a vast amount of training data to ensure accurate predictions in all scenarios. As a consequence, there is a need to label every dynamic and static object with 3D bounding boxes and additional attributes over hundreds or thousands of hours of driving. However, manually creating these labels is expensive, time-consuming, and error-prone. While several datasets with 3D bounding box annotations are available for dynamic objects \cite{sun2020scalability}, \cite{caesar2020nuscenes}, \cite{geiger2012we}, \cite{matuszka2023aimotive}, the number of available static object datasets with 3D annotations \cite{fent2024man}, especially those containing distant objects, is remarkably limited. As a result, there is a significant interest in automating the generation of such training data without human intervention. Our primary goal is to provide accurate 3D bounding boxes for traffic management objects, ensuring that the projected 2D bounding boxes in the camera image encompass objects from a wide range of viewing angles and distances. This step is crucial for all downstream tasks of the proposed method, such as classification or optical character recognition. Since the data recording process typically involves multiple sensors and a high frame rate, this requirement is easily met.

The main contribution of this work is a novel method that provides accurate positioning with an average mean distance of 0.2-0.3 meters and temporally consistent 3D bounding boxes of traffic management objects up to 200 meters away. Our method also determines additional attributes such as traffic light state, traffic light mask type, traffic sign type, and occlusion. The proposed solution is simple yet effective, relying solely on 2D images and Global Navigation Satellite System/Inertial Navigation System (GNSS/INS) data, without the need for expensive active sensors like LiDAR. Furthermore, we publish a representative dataset, automatically generated using our algorithm, under a CC BY-NC-SA 4.0 license, allowing the research community to use it for non-commercial research purposes\footnote{\href{https://github.com/aimotive/aimotive_tl_ts_dataset}{https://github.com/aimotive/aimotive\_tl\_ts\_dataset}}. To our knowledge, no publicly available large-scale dataset including distant objects currently exists that contains accurate 3D bounding boxes of traffic management objects, particularly traffic lights.


\section{Related Work}
\label{sec:related_work}

Automatic 3D localization methods for static objects, particularly traffic signs, are already available with certain limitations. The three main approaches are the following: 1) using LiDAR point cloud data to identify the cluster associated with the object; 2) generating a synthetic point cloud through Structure-from-Motion and associating 2D image-space detections to the resulting 3D points; and 3) applying triangulation using camera images, GNSS, and orientation information.

Approach 1) is well-suited for traffic signs due to their highly reflective coating, which produces dense point groups in LiDAR data with high-intensity values that can be effectively clustered. Soil\'{a}n et al. in \cite{soilan2016} used this technique to localize traffic signs, reprojecting them onto 2D camera images to spatially and temporally synchronize with the point cloud data. While this method can yield accurate results, separating traffic signs close to each other is challenging. Another drawback, as they noted, is that in urban environments, the rate of false positive detections increases due to the higher number of reflective objects. A similar approach \cite{ghallabi2019} was presented by Ghallabi et al., but in their case, no camera information was used and the method was only tested in a highway environment. Song and Myung described a method in \cite{camera_lidar_tsr} that also utilizes 2D image detection and LiDAR point cloud data. They first apply a deep learning model to camera images to predict 2D bounding boxes of traffic signs. These boxes are then used to filter relevant parts of the point cloud within a frustum, and DBSCAN clustering is applied to eliminate non-relevant point groups. However, this group of work depends heavily on the quality of the point cloud. For traffic signs located far from the observer or higher than the LiDAR detection range, few or no reflective points are detected, leading to low localization accuracy and an increased number of false negative detections. Additionally, this method is ineffective for traffic lights, as they are mostly black and have lower reflectivity. Moreover, most traffic lights are positioned higher than the detection range of LiDAR sensors.

Approach 2) is primarily used to create large-scale but low-resolution maps of traffic signs. Structure-from-Motion relies on identifying features in consecutive camera images, associating them, and estimating their 3D position through triangulation, thereby generating a synthetic point cloud from the images. Musa's solution \cite{musa2022} is based on this method and further improves localization accuracy using the GNSS coordinates of the images. Although the algorithm runs in real-time, its accuracy is around 2.75 meters, which is insufficient for automated ground truth data generation. Mapillary\footnote{\url{https://www.mapillary.com}} provides a world-scale map of traffic management objects using dashcam images and Structure-from-Motion. However, based on our experiments, the accuracy is also within several meters, and only latitude/longitude positions can be downloaded. No 3D bounding boxes are available that could be projected onto camera images. Therefore, this solution cannot be used for automated ground truth generation either.

The last group of methods relies on image-space detections, GNSS, and orientation information. Mentasti et al. developed a localization algorithm \cite{mentasti2023} for traffic lights, which they applied to the DriveU Traffic Light Dataset \cite{driveu}. They estimated individual distances of traffic lights for each 2D detection using disparity maps, applied a tracking algorithm, and finally averaged the positions for each track ID. However, the 3D position estimation was not validated since the DriveU dataset only provides 2D bounding boxes of traffic lights. Fairfield and Urmson used a traffic light detection algorithm \cite{fairfield2011} that identifies brightly colored red, amber, and green blobs in the image. These detections are then associated between frames using image-to-image association and least squares triangulation. The orientation of the traffic light is estimated as the reciprocal heading of the mean car heading over all the image labels used to estimate the traffic light position. In traffic light online detection, the map positions are projected into the image plane, and a region of interest is defined, considering a larger area than the predicted bounding box. Finally, the classifier is applied to the image cutouts to find the light blobs and classify the colors. Since disparity-based depth estimation is known to be inaccurate in long distances and color-based blob detection is not applicable in the case of traffic signs, these methods cannot be applied to accurate 3D automatic annotation of traffic lights and signs.

To summarize, there is currently no comprehensive algorithm for automatically generating high-precision 3D bounding boxes (including distant objects) of traffic signs and lights with additional attributes. The existence of such an algorithm could have a significant impact on the development of image-based neural networks used by self-driving vehicles.

\section{Automatic Annotation of Traffic Lights and Signs in 3D}
\label{sec:method}

\begin{figure}
\centering
\includegraphics[width=1.0\linewidth]{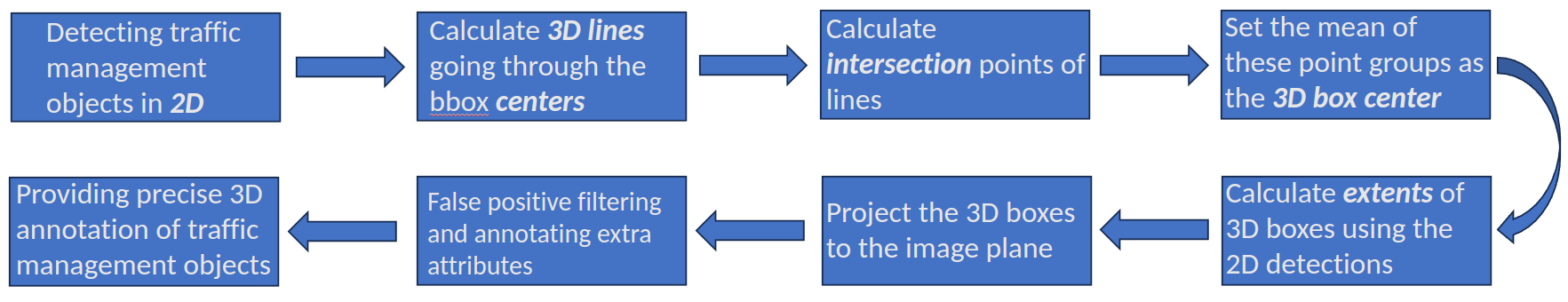}
\caption{The main steps of the automatic annotation method.}
\label{fig:flowchart}
\end{figure}

Our proposed method, depicted by Figure \ref{fig:flowchart}, can be used for generating unlimited amounts of 3D training data for traffic management objects. This automatic annotation algorithm consists of five steps: 1) Mask2Former \cite{cheng2021mask2former} image segmentation model is used to obtain the 2D positions of traffic lights and traffic signs; 2) 3D bounding box centers are localized by triangulating the lines of sight in the Earth-centered, Earth-fixed coordinate system (ECEF), resulting in a 3D map of traffic management objects; 3) 3D bounding box extent and orientation are estimated; 4) 3D boxes are transformed into the instantaneous coordinate systems (i.e., vehicle coordinate system) of each frame; and 5) 3D boxes are projected onto the camera image plane and 2D image cutouts of traffic management objects are classified. The outcome of the proposed method is a dataset containing 3D annotations of traffic lights and traffic signs for each frame, including information on color state, occlusion, traffic light mask type, and traffic sign type. We describe the details of the main steps of our method in the following subsections.

\subsection{3D Localization}

The first step in 3D localization involves acquiring 2D detections of traffic management objects in images captured by a single front camera. Then, the bounding boxes are calculated and the centers of the bounding boxes are stored. Only predicted 2D bounding boxes with high confidence are used, thereby excluding false positive detections. This step does not reduce the recall of 3D detection, as traffic management objects will typically be close to the ego vehicle's trajectory during recording and will appear large enough in the images over a sufficient time horizon to ensure highly confident 2D predictions.

\begin{figure}
\centering
\includegraphics[height=4.5cm]{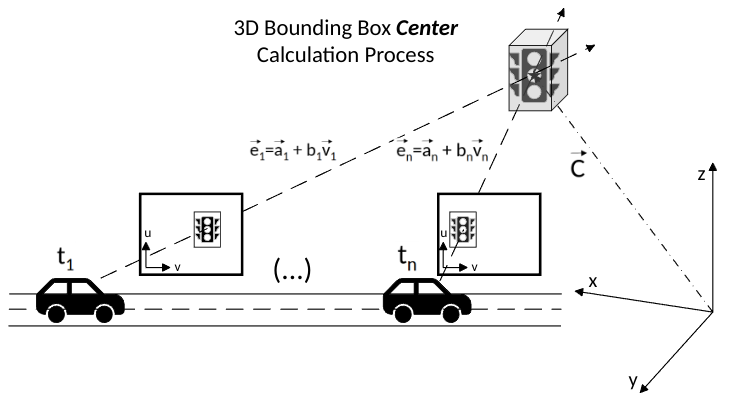}
\caption{Calculation of 3D bounding box center.}
\label{fig:3D_center}
\end{figure}

The next step is to calculate the 3D positions of these static objects. To apply the triangulation technique, 2D observations of the same physical 3D point from multiple viewing angles are needed. Since traffic lights are relatively small and compact objects and traffic signs are planar, the center of the 2D bounding box can be treated as the projection of the same physical point with good approximation. Using the GNSS and orientation data of the observer along the ego vehicle's trajectory, as well as the 3D lines pointing towards the 2D bounding box centers, 3D positions of the object center in a global coordinate system through the triangulation technique illustrated in Figure \ref{fig:3D_center} are determined.

Specifically, 3D lines that come closer than 10 centimeters to each other are collected. Then, the coordinates of the point closest to the lines are calculated by iterating over these line pairs. This process generates many candidate points for the centers of 3D boxes, which are then aggregated using the DBSCAN clustering method \cite{ester1996density}. A 3D point forms a cluster if there are at least 3-5 points within 5-10 centimeters of each other. After identifying these clusters, their average is taken as the final prediction of the 3D box center in ECEF coordinates. The distance filtering and clustering steps enhance the algorithm's robustness against random errors related to GNSS position, orientation, or camera calibration. It's important to note that this method does not require object tracking, as localization is calculated directly in the global coordinate system. This leverages the fact that the likelihood of incorrectly associating two 2D detections from different physical objects in 3D space, given such low distance threshold values in the triangulation process, is very low.

\subsection{Extent Calculation}

\begin{figure}
\centering
\includegraphics[height=5.5cm]{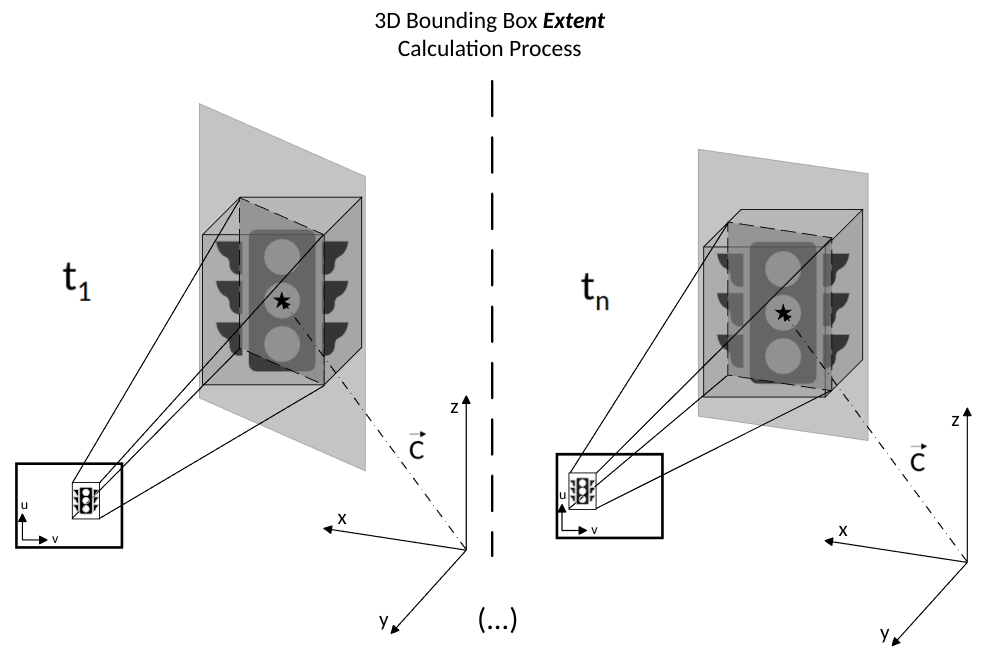}
\caption{Calculation of 3D bounding box extent.}
\label{fig:extent}
\end{figure}

The map with the bounding box centers of traffic management objects is provided after the localization step. However, the extent of the detected objects is still unknown. To determine this attribute of traffic lights, the intersections of the lines pointing towards the 2D bounding box corners with a vertically aligned plane that contains the center of the object and is perpendicular to our line of sight in the x-y plane are calculated. In this step, the cross-sections of the 3D bounding boxes from various viewing angles are measured. Finally, the widths and heights of these cross-sections are averaged to estimate the width, depth, and height of the 3D bounding boxes. Note that the width and depth are set to the same value, which is a good estimate for the commonly vertically aligned traffic lights. The visualization of the traffic light size estimation method is illustrated in  Figure \ref{fig:extent}. 

Traffic signs have a larger variety of shapes and can appear in shapes other than rectangles (e.g., circles, triangles). Therefore, instead of using the corners of the 2D bounding boxes, the intersections of the vertical plane and the lines pointing toward the edge points of the bounding box are calculated. Since traffic signs are planar objects, the maximum of the measured widths are taken and the depth is set to 10 centimeters.

\subsection{Orientation Estimation}

\begin{figure}
\centering
\includegraphics[height=3.5cm]{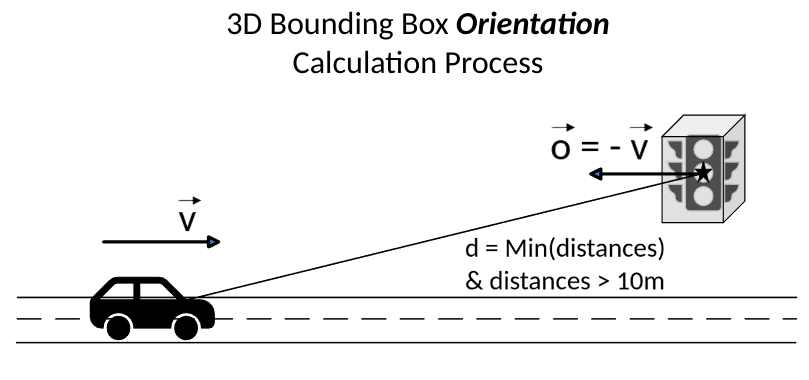}
\caption{Calculation of 3D bounding box orientation.}
\label{fig:orientation}
\end{figure}

Our proposed algorithm employs a heuristic approach to determine the orientation of traffic lights. The orientation estimation method identifies the frame where the vehicle is approximately 10 meters in front of the traffic light and assumes it is oriented opposite to the direction of travel. While this method generally provides accurate orientations for relevant traffic lights, it may be incorrect for cross-traffic ones. However, this does not affect the generation of 2D image cutouts for classification tasks, as the 2D projection of vertically aligned traffic light boxes remains relatively consistent regardless of different rotation angles around the Z axis (see Fig. \ref{fig:orientation}). 

For traffic signs, the algorithm uses the line-of-sight vector to the road sign in the frame where the measured width is maximal. The final orientation is the reverse of this vector, indicating the vehicle was closest to being directly opposite the corresponding traffic sign.

\subsection{Reducing False Positive Detections}

At this stage, a map of 3D bounding boxes for traffic management objects with high positional accuracy (within 0.2-0.3 meters from the ground truth, see details in Section \ref{sec:eval}) is created, which can be used in various operational design domains such as rain, night, snow, etc. From this map, we generate 2D image cutouts of traffic management objects by projecting them onto the camera image plane, up to 200 meters from the ego vehicle position. Based on our experience, measurement errors in the triangulation technique can produce false positive boxes that are located on the same 3D lines as the true positive box. These false positives can be eliminated by associating their 2D projections with the original 2D bounding boxes. During this process, we first calculate the intersection-over-union (IoU) between the projections and the 2D bounding boxes, associating the average IoU value over the frames for each 3D bounding box. We then group 3D boxes that appear very close to each other, defined by an angle between their line of sight vectors below 0.25-0.3 degrees across several camera frames. Finally, we select the 3D box with the highest IoU value from each group as the final prediction.

\subsection{Classification of Object Attributes}

When considering the attributes of traffic management objects, we differentiate between time-dependent and time-independent properties. Time-dependent attributes, such as the traffic light color or the occlusion of traffic management objects, must be classified for each frame, which can be challenging when the object is far away from the ego vehicle. In contrast, time-independent attributes, such as the types of objects (e.g., forward arrow traffic light or yield, stop sign), do not change over time. Therefore, we can use high-resolution image cutouts when the ego vehicle is close to the objects. To automatically classify these attributes, we utilize standard convolutional neural networks.

\begin{figure*}
  \begin{subfigure}{0.5\linewidth}
  \includegraphics[width=0.8\linewidth]{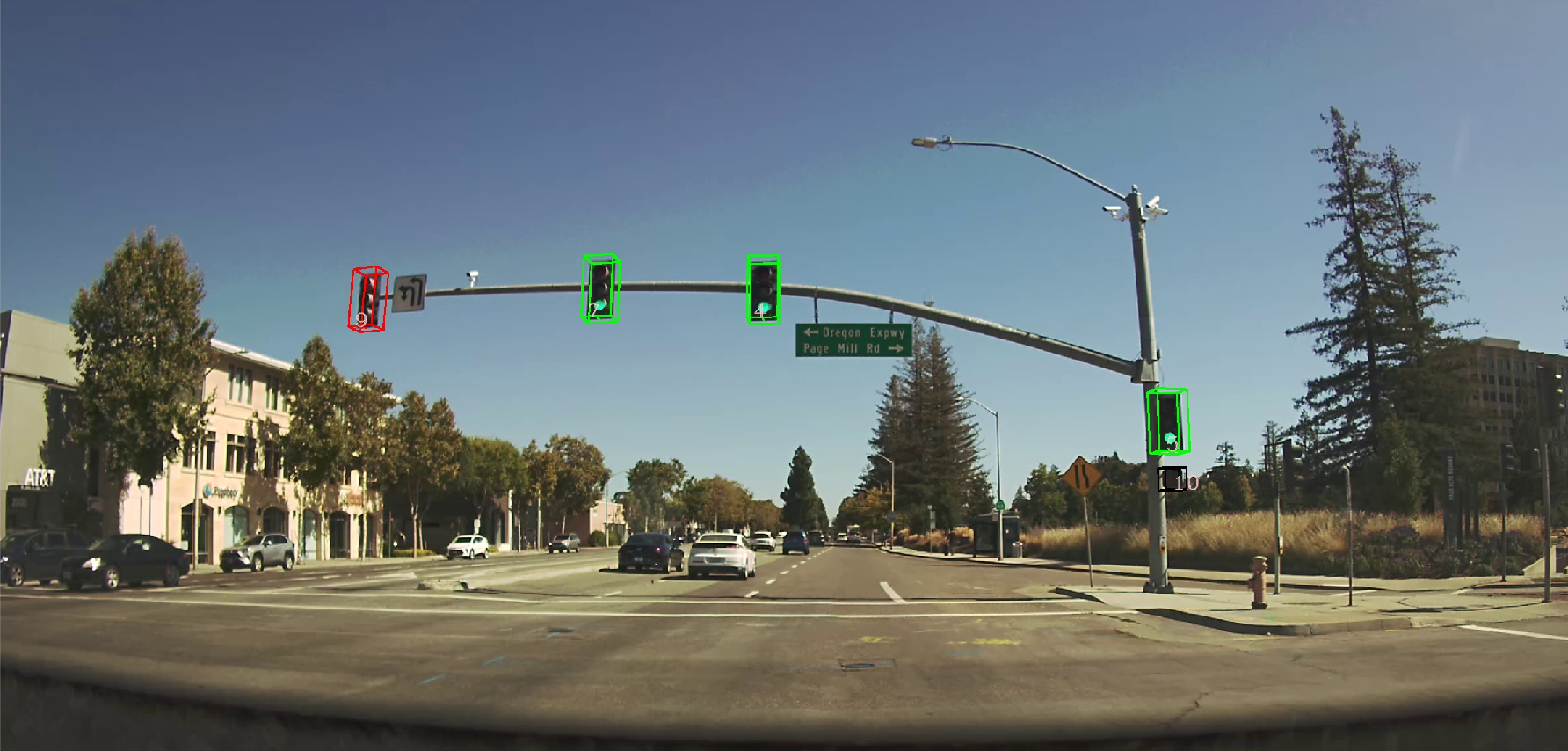}
  \end{subfigure}
  \hfill
  \begin{subfigure}{0.5\linewidth}
  \includegraphics[width=0.93\linewidth]{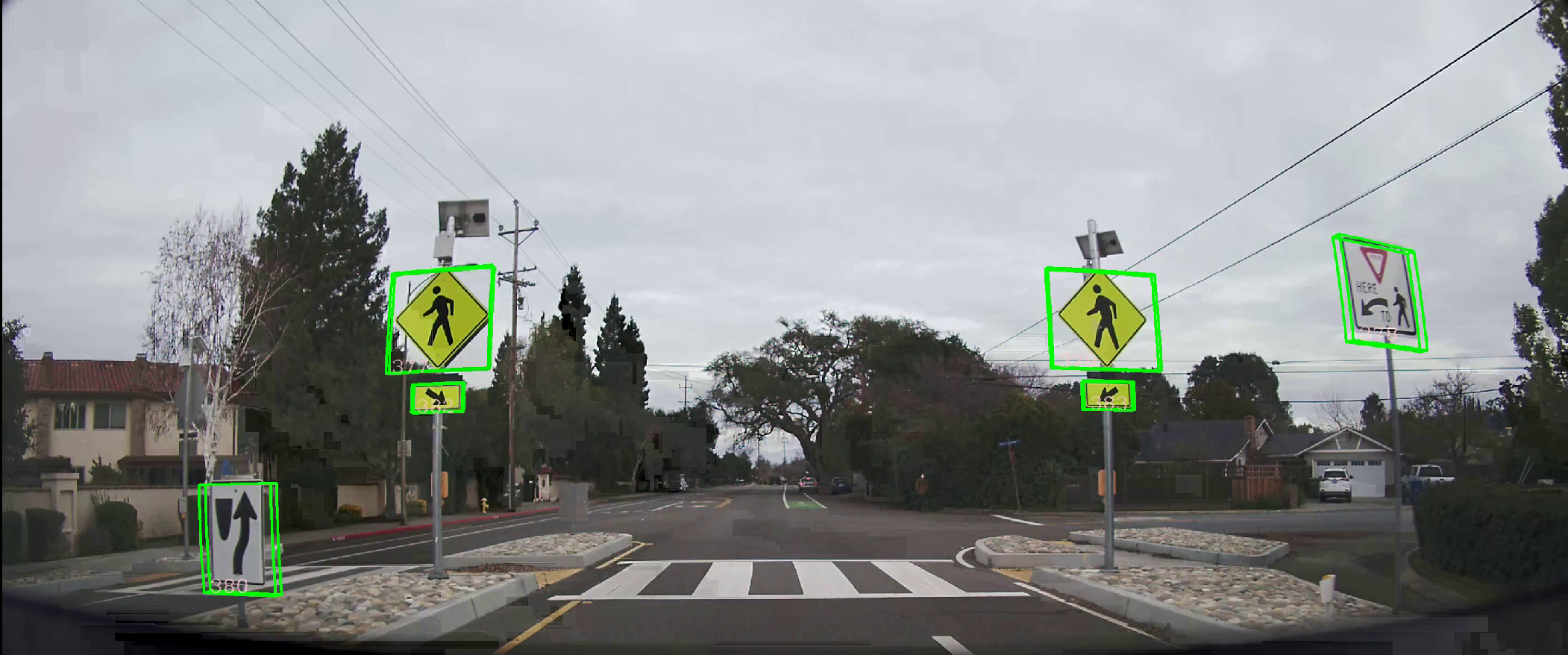}
  \end{subfigure}
  \caption{Samples from the dataset with 3D traffic sign and light annotations. The bounding boxes are automatically generated by our method. Traffic light states are color-coded.}
  \label{fig:tl_ts_autoannot_vis}
\end{figure*}

\section{3D Traffic Light and Road Sign Dataset}

\begin{figure*}
  \begin{subfigure}{0.5\linewidth}
  \includegraphics[width=0.8\linewidth]{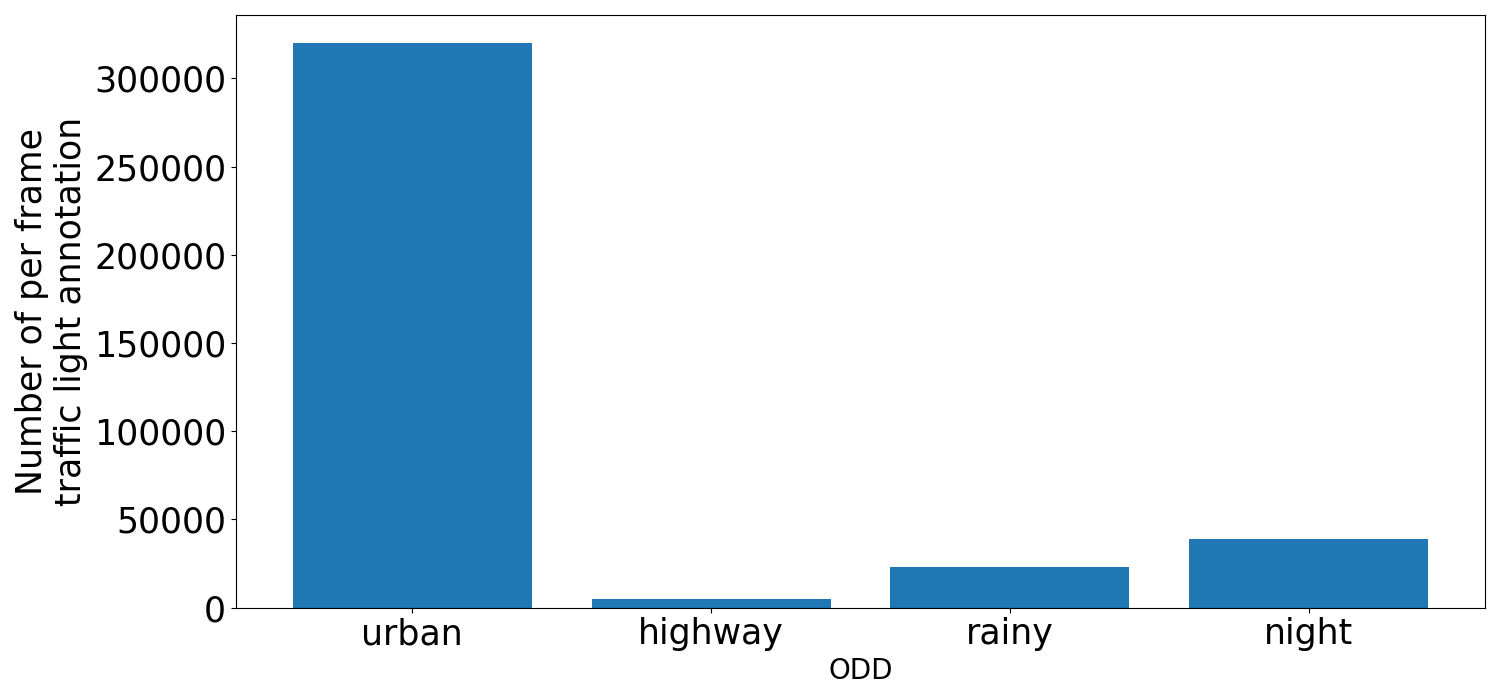}
  \end{subfigure}
  \hfill
  \begin{subfigure}{0.5\linewidth}
  \includegraphics[width=0.93\linewidth]{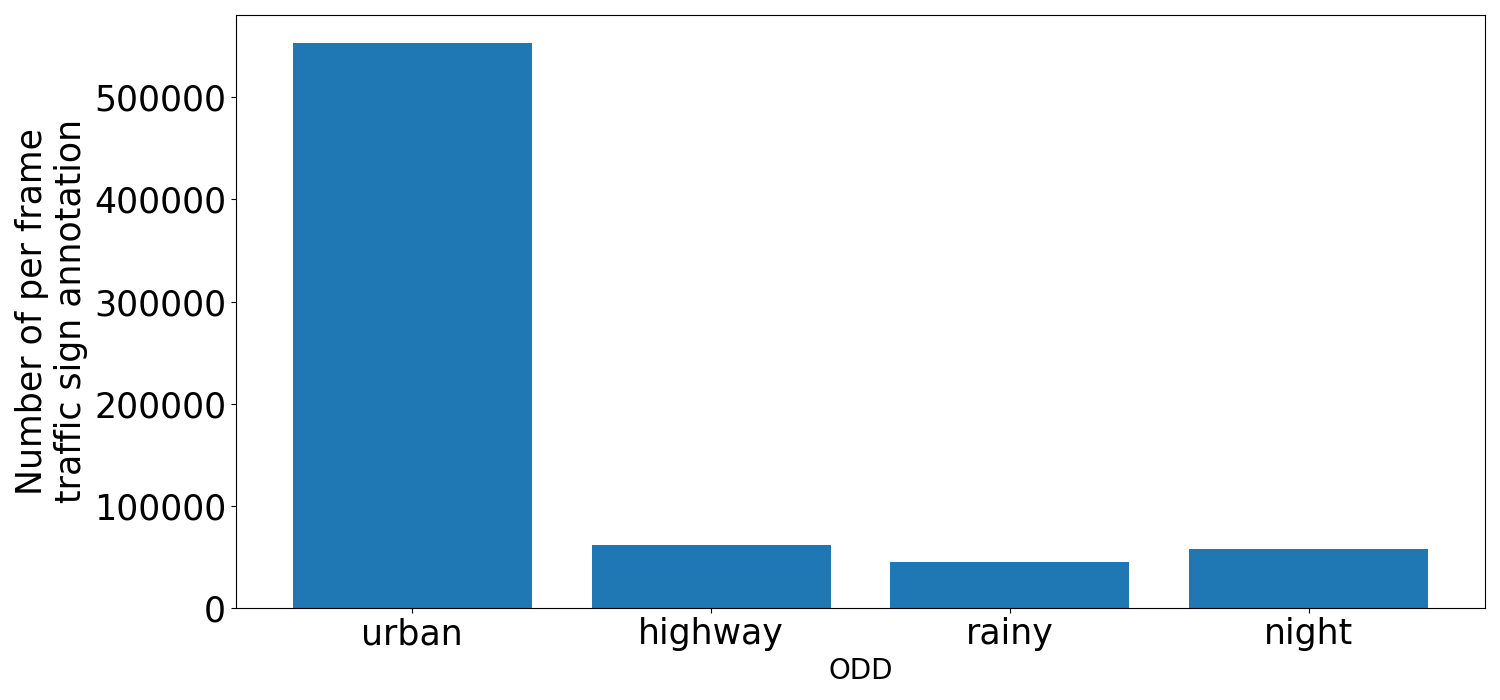}
  \end{subfigure}
  \caption{Data distribution of per frame annotations.}
  \label{fig:per_frame_annot}
\end{figure*}

\begin{figure}
\centering
\includegraphics[width=0.6\linewidth]{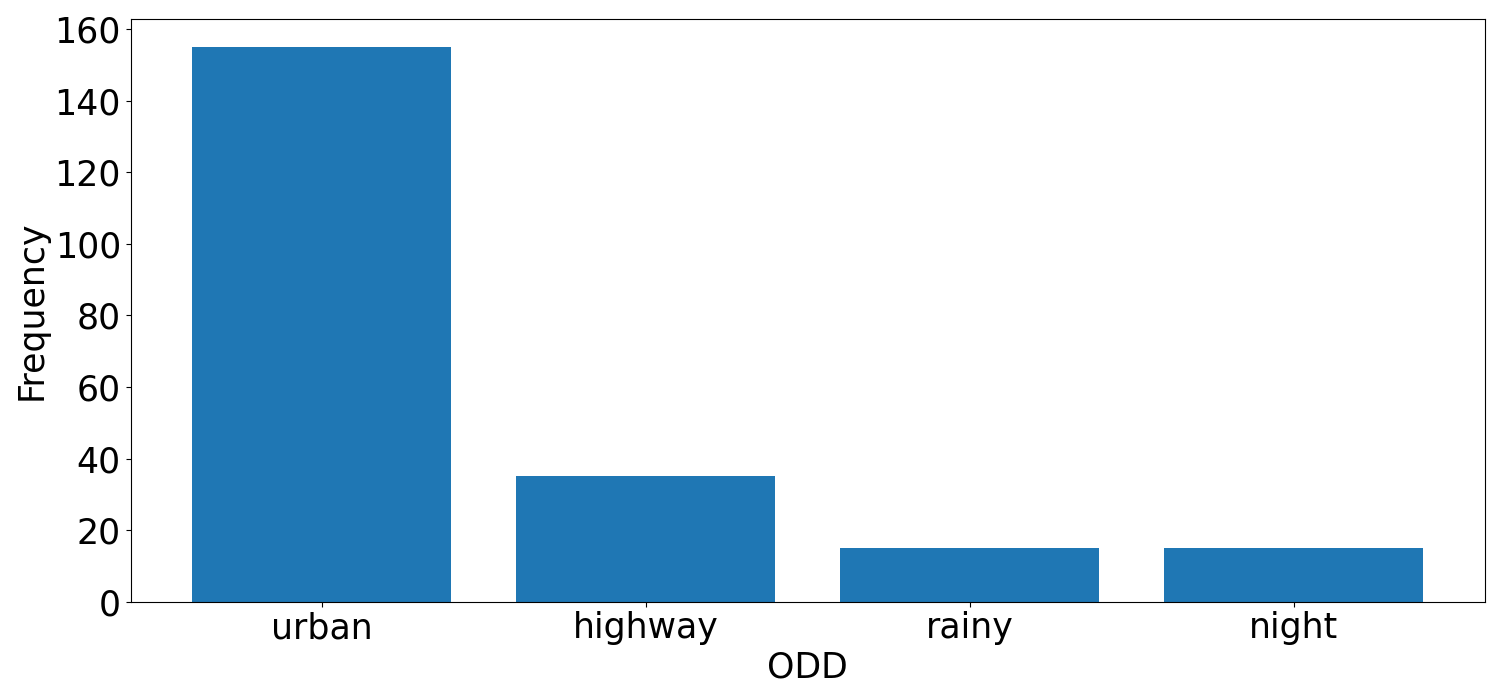}
\caption{Data distribution across the different operational design domains.}
\label{fig:data_dist}
\end{figure}

To facilitate research in static 3D object detection and address the challenges mentioned in Section \ref{sec:intro}, we have published a diverse training dataset of traffic lights and road signs, generated by our method described in Section \ref{sec:method}. The recordings were captured in two countries (California, US, and Hungary) in urban and highway environments, and under different times of day and weather conditions. The dataset includes approximately 50,000 3D auto-annotated frames from 220 sequences, each 15 seconds long, totaling 55 minutes of driving. Figure \ref{fig:tl_ts_autoannot_vis} visualizes sample annotations of the dataset. The sequences consist of images captured by four different cameras: wide and narrow front cameras, as well as left and right cross-traffic cameras. Each frame includes a JSON annotation file for the traffic light and traffic sign 3D bounding boxes, which provides geometric information along with the traffic light state and mask, traffic sign type, object occlusion, and the text on traffic signs (extracted using the Google Vision API). The data distribution across the ODDs is shown in Figure \ref{fig:data_dist}. The majority of the dataset consists of urban scenes, with approximately 320,000 auto-annotated traffic lights and 550,000 traffic signs. The per-frame annotation distribution is depicted in Figure \ref{fig:per_frame_annot}.

\section{Evaluation}
\label{sec:eval}

\subsection{Validation Challenges}

Precise localization of traffic management objects on a large scale is extremely challenging due to issues such as sensor limitations described in Section \ref{sec:related_work}. This challenge explains why there is still no publicly available dataset with long-range 3D annotations for traffic signs and traffic lights. Although Mapillary provides global latitude and longitude coordinates for traffic signs, the accuracy is low, and there is no information about the vertical position, extent, or orientation to accurately place these objects in the local coordinate system of a driving scene. Popular autonomous driving datasets like nuScenes, KITTI, and Waymo present additional challenges. Among these, only Waymo \cite{waymo} provides 3D bounding boxes for traffic signs and has GNSS information for the camera frames, which is necessary to evaluate our algorithm on a dataset. However, it contains annotations up to only 77-78 meters from the observer, and there is no information about the relevance of the traffic sign to the ego vehicle, hence we cannot directly measure precision or recall, but only a distance error between the associated ground truth-prediction pairs. Moreover, we are unaware of publicly available traffic light datasets with 3D annotations, especially those containing distant objects. Given these difficulties, we have decided to validate our algorithm not only on the Waymo traffic sign dataset but also using manually annotated in-house benchmark datasets.

\begin{figure*}
  \begin{subfigure}{0.5\linewidth}
  \includegraphics[width=0.93\linewidth]{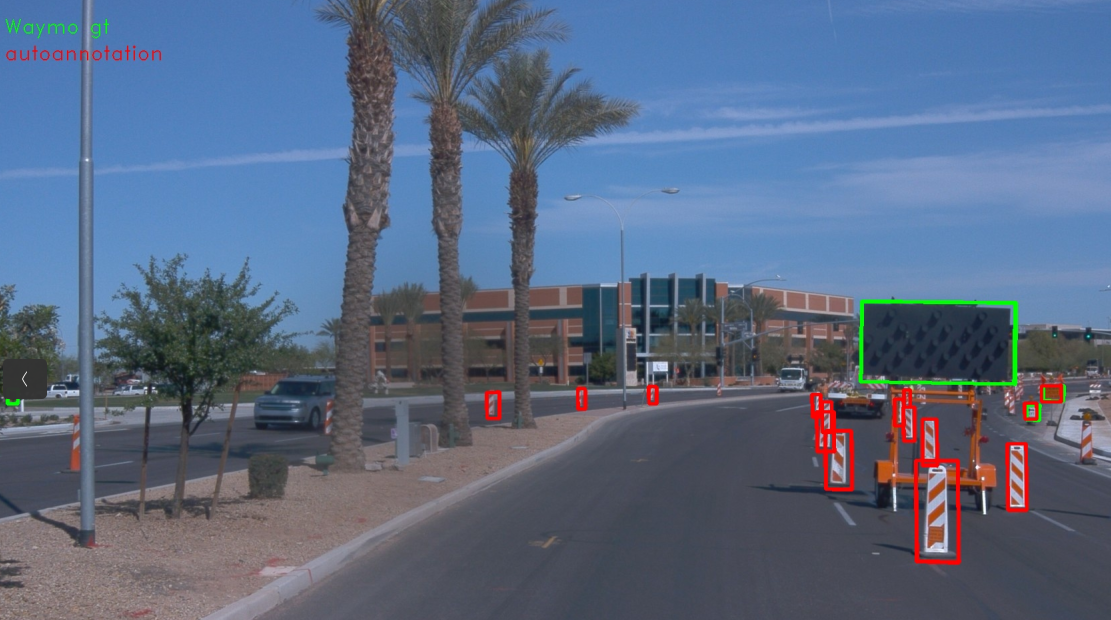}
  \end{subfigure}
  \hfill
  \begin{subfigure}{0.5\linewidth}
  \includegraphics[width=0.93\linewidth]{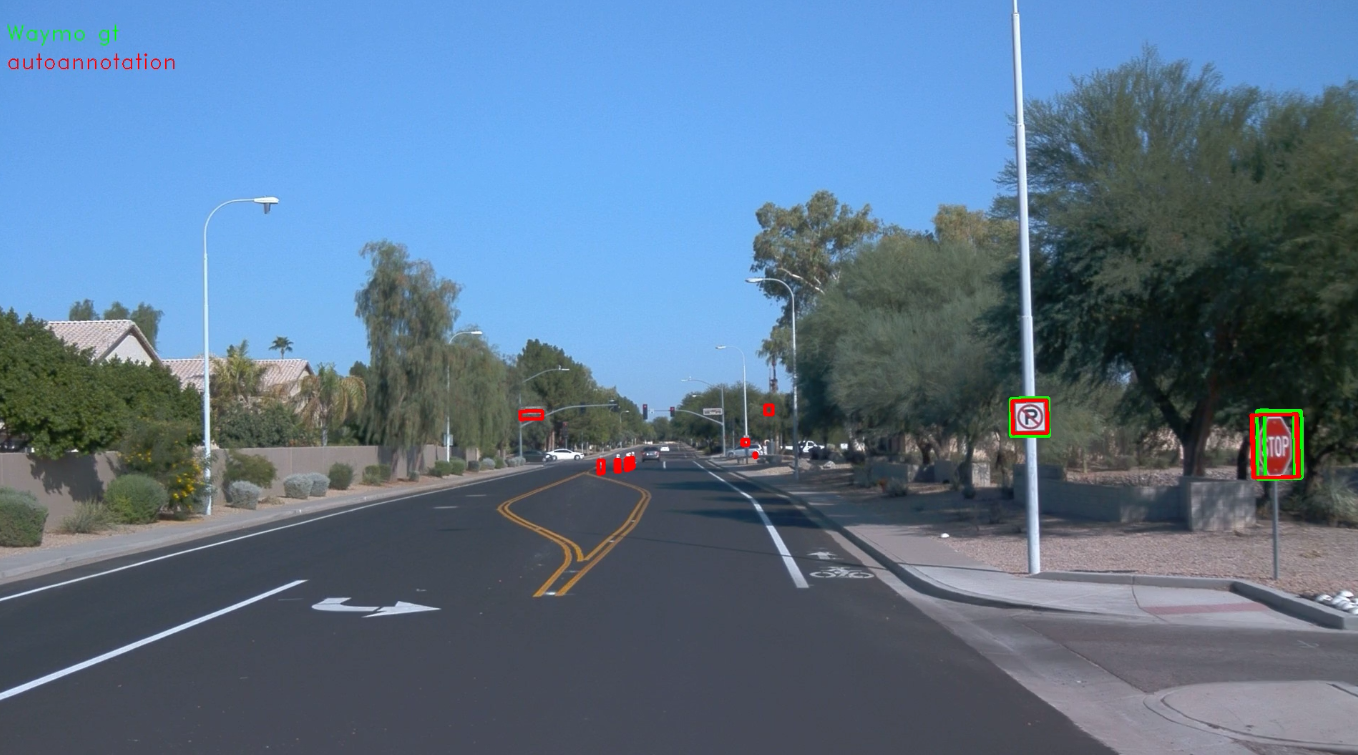}
  \end{subfigure}
  \caption{Qualitative comparison of Waymo ground truth (green) and auto-annotated (red) 3D bounding boxes (\textbf{Left}: segment-14811410906788672189\_373\_113\_393\_113\_with\_camera\_labels;  \textbf{Right}: segment-10203656353524179475\_7625\_000\_7645\_000\_with\_camera\_labels). The annotated traffic sign types in the ground truth and in the automatic annotation can be very different.}
  \label{fig:waymo_comp}
\end{figure*}

\begin{figure}
    \centering
  \begin{subfigure}{0.49\columnwidth}
  \includegraphics[width=\linewidth]{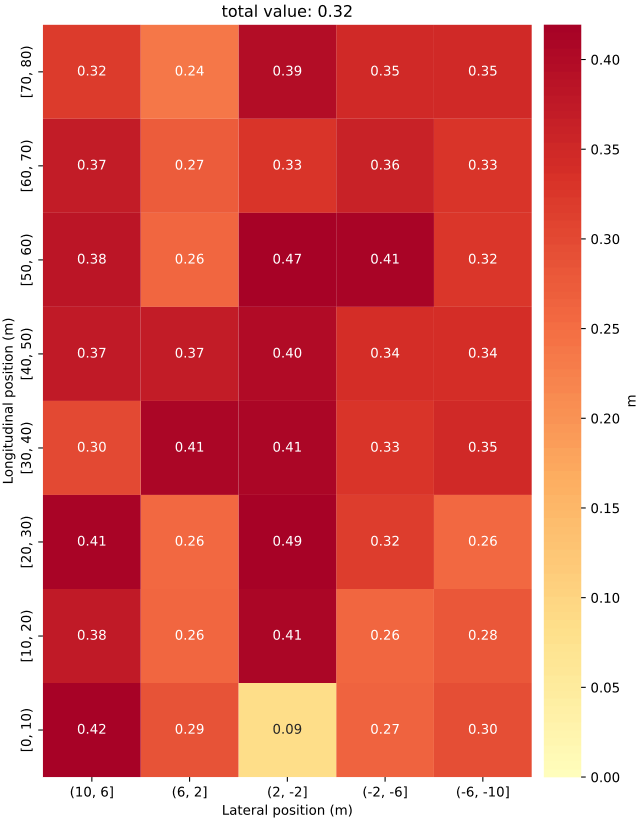}
  \end{subfigure}
  \hfill
  \begin{subfigure}{0.49\columnwidth}
  \includegraphics[width=\linewidth]{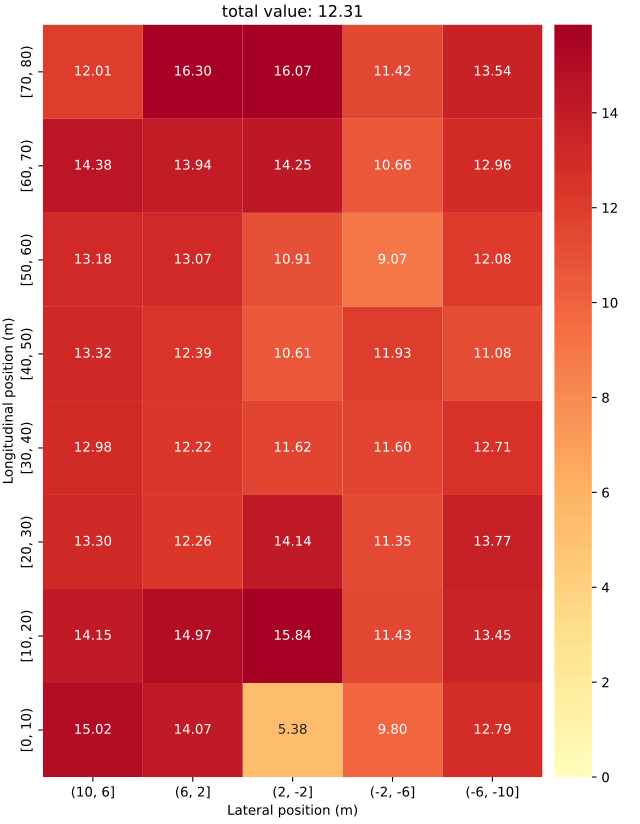}
  \end{subfigure}
  \caption{Evaluation results (in 4 m x 10 m blocks until 80 meters) of the proposed algorithm measured on \textbf{all} traffic sign boxes related to the Waymo validation set (\textbf{Left}: Mean error in bounding box center estimation (\textbf{0.32 meters}). \textbf{Right}: Mean absolute error in box orientation (\textbf{12.31 degrees}). (best viewed by zooming in)}
  \label{fig:waymo_metrics_all}
\end{figure}

\subsection{Validation of the method on the Waymo dataset}

We evaluated our proposed method on the validation set of Waymo. Since our algorithm relies on egomotion-based triangulation, we filtered out segments where the traveled distance was less than 3 meters. Hence, we ended up with a final validation set containing 189 segments. For the comparison of all detected traffic signs with all Waymo ground-truth boxes, we omitted classification metrics such as precision/recall due to the different definitions of the classes between Waymo and Mask2Former which we used to determine the existence of traffic signs in images. Fig. \ref{fig:waymo_comp} depicts an example of the class definition mismatch. Our algorithm provides 3D bounding boxes only for traffic signs detected by the Mask2Former model. All metrics were calculated within the range of [-10m, 10m] lateral and [0m, 80m] longitudinal positions of the instantaneous coordinate system. The association distance threshold was set to 1 meter. Altogether \textbf{45,257} Waymo ground truth boxes have been associated with the bounding boxes generated by our method. The absolute mean distance between the centers is \textbf{0.32 $\pm$ 0.22 meters} and the mean absolute difference in the orientation is \textbf{12.31 degrees} (see metrics in Table \ref{tab:ts_validation_results_waymo}). The error distributions are shown in Fig. \ref{fig:waymo_metrics_all}, where the performance was evaluated in 4 m x 10 m blocks.

\begin{table}[t]
\begin{center}
\caption{Quantitative evaluation results of our automatic annotation method for \textbf{all} traffic signs of the Waymo validation dataset.}
\label{tab:ts_validation_results_waymo}
\begin{tabular}{ll}
\hline\noalign{\smallskip}
Metric & Result\\
\noalign{\smallskip}
\hline
\noalign{\smallskip}
Localization error  & 0.32 $\pm$ 0.22 meters\\
Orientation error  &  12.31 degrees\\
\hline
\end{tabular}
\end{center}
\end{table}

\begin{figure}
    \centering
  \begin{subfigure}{0.49\columnwidth}
  \includegraphics[width=\linewidth]{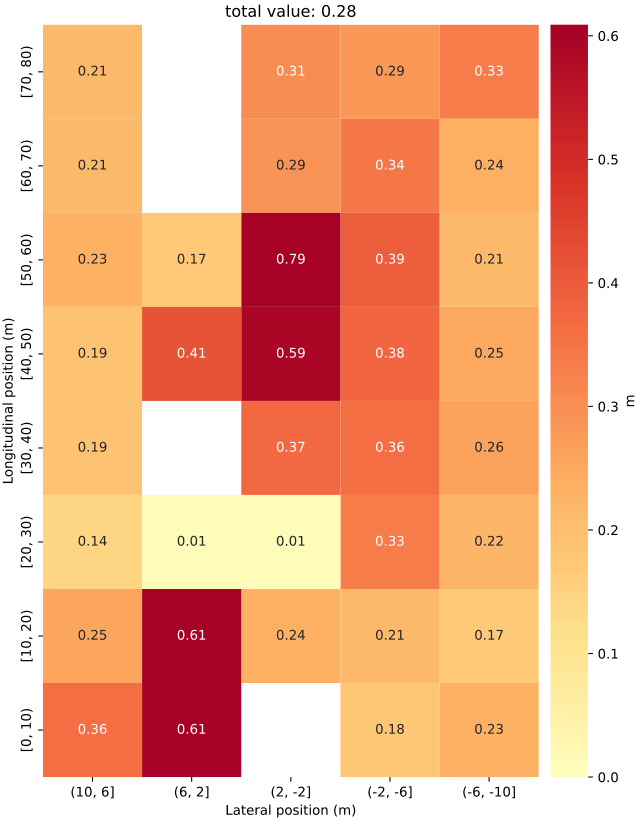}
  \end{subfigure}
  \hfill
  \begin{subfigure}{0.49\columnwidth}
  \includegraphics[width=\linewidth]{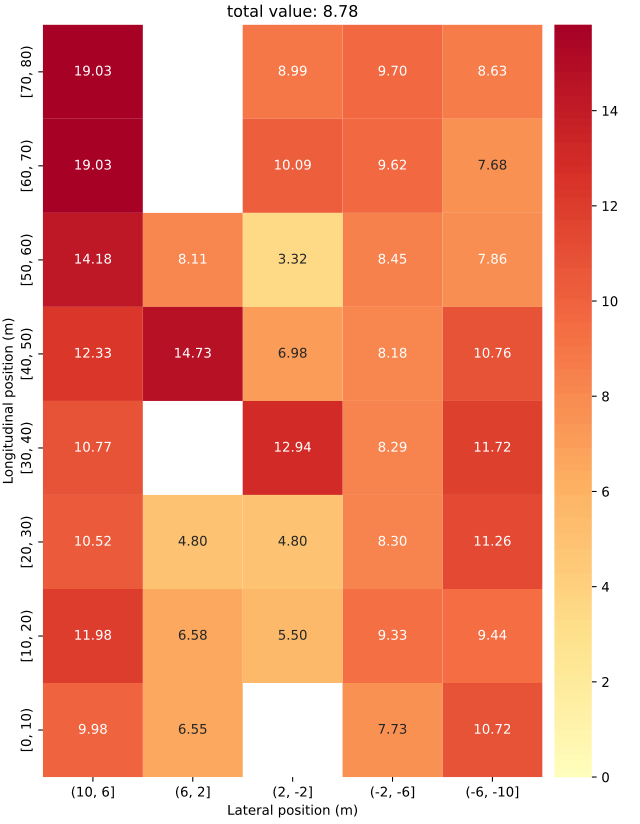}
  \end{subfigure}
  \caption{Evaluation results (in 4 m x 10 m blocks until 80 meters) of the proposed algorithm measured on the manually selected \textbf{speed limit and stop} signs related to the Waymo validation set. \textbf{Left}: Mean error in bounding box center estimation (\textbf{0.28 meters}). \textbf{Right}: Mean absolute error in box orientation (\textbf{8.78 degrees}).}
  \label{fig:waymo_metrics_relevant}
\end{figure}

\begin{table}[t]
\begin{center}
\caption{Quantitative evaluation results of our automatic annotation method for \textbf{speed limit and stop signs} of the Waymo validation dataset.}
\label{tab:ts_validation_results_waymo_relevant}
\begin{tabular}{ll}
\hline\noalign{\smallskip}
Metric & Result\\
\noalign{\smallskip}
\hline
\noalign{\smallskip}
Recall  & 93.76 \%\\
Localization error & 0.28 $\pm$ 0.23 meters\\
Orientation error  &  8.78 degrees\\
\hline
\end{tabular}
\end{center}
\end{table}

We also provide validation results with respect to a relevant subset of traffic signs where we manually selected speed limit and stop signs from the mentioned 189 segments. In case of four traffic signs we did not approach them closer than 40 meters during the segment, and therefore our algorithm could not provide reliable bounding box estimation. Ignoring these objects, we measured the recall, position, and orientation error on 66 physically different traffic signs. Together, \textbf{5,511} ground truth boxes have been associated with our detections, where we detected \textbf{93.76 \%} of traffic signs. The absolute mean distance between the centers is \textbf{0.28 $\pm$ 0.23} meters and the mean absolute difference in orientation is \textbf{8.78 degrees} (see Table \ref{tab:ts_validation_results_waymo_relevant}). Detailed metrics can be seen in Fig. \ref{fig:waymo_metrics_relevant}. These results indicate that the performance of our algorithm is even better if we consider only the traffic signs that are critical for self-driving.

\subsection{Validation of Automatic Traffic Sign Annotation on in-house dataset}

\begin{figure}
\centering
\includegraphics[height=4.5cm]{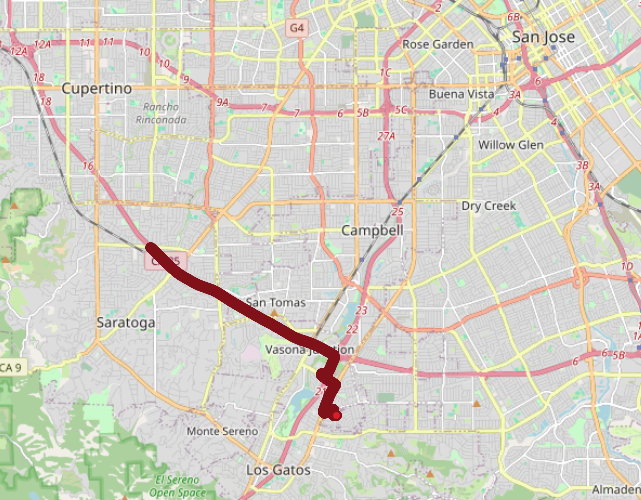}
\caption{Visualization of the traffic sign validation route.}
\label{fig:ts_validation_route}
\end{figure}

We also validated the traffic sign automatic annotation performance on a 7-kilometer route in San José, California, which included both highway and urban sections (see the validation route in Figure \ref{fig:ts_validation_route}). In total, 183 traffic signs were manually annotated with oriented 3D bounding boxes using LiDAR point cloud data. This manually created map was projected into the instantaneous coordinate systems of the vehicle, allowing for a detailed comparison with the automatic annotation. All metrics were calculated within the range of [-10m, 10m] lateral and [0m, 200m] longitudinal positions of the instantaneous coordinate system. The association distance threshold was set to 1 meter, and we calculated localization precision and recall related to the bounding box center. The automatic annotation method achieved \textbf{97.08\%} precision and \textbf{95.33\%} recall (see Table \ref{tab:ts_validation_results} for more detailed results). It is worth noting that the lower recall value resulted from only six missed traffic signs on the highway section, which included traffic signs with categories less relevant for self-driving (e.g. destination distance, interchange advance exit).

\begin{figure}
    \centering
  \begin{subfigure}{0.49\columnwidth}
  \includegraphics[width=\linewidth]{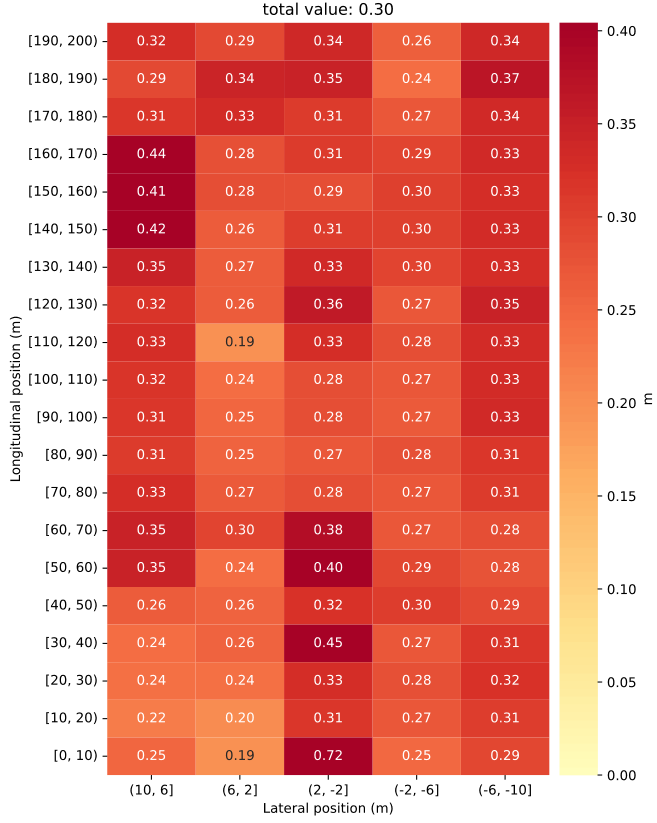}
  \end{subfigure}
  \hfill
  \begin{subfigure}{0.49\columnwidth}
  \includegraphics[width=\linewidth]{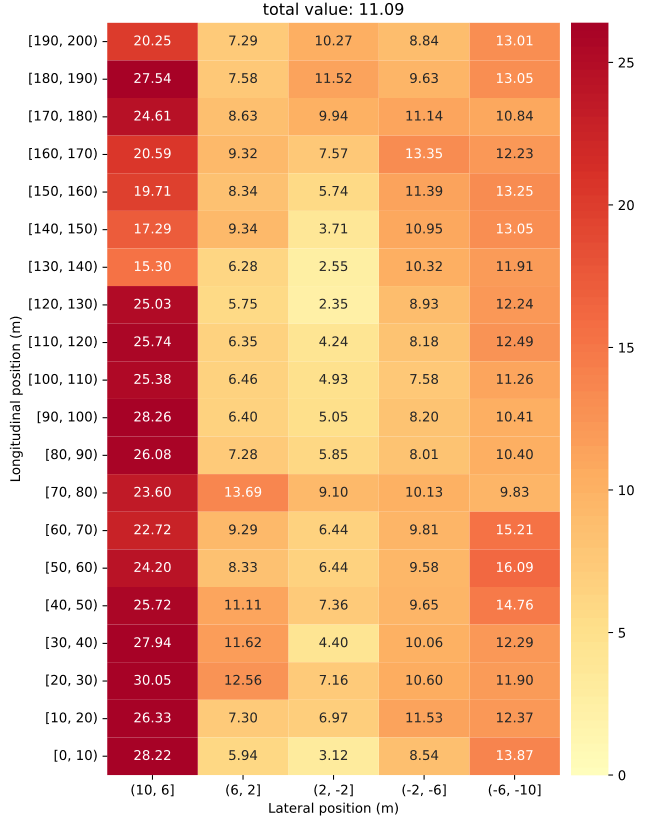}
  \end{subfigure}
  \caption{Evaluation results (in 4 m x 10 m blocks until 200 meters) of the proposed algorithm measured on our manually annotated \textbf{in-house} traffic sign dataset. \textbf{Left}: Mean error in bounding box center estimation (\textbf{0.3 meters}). \textbf{Right}: Mean absolute error in box orientation (\textbf{11.09 degrees}).}
  \label{fig:ts_metrics}
\end{figure}

\begin{figure}
    \centering
  \begin{subfigure}{0.49\columnwidth}
  \includegraphics[width=\linewidth]{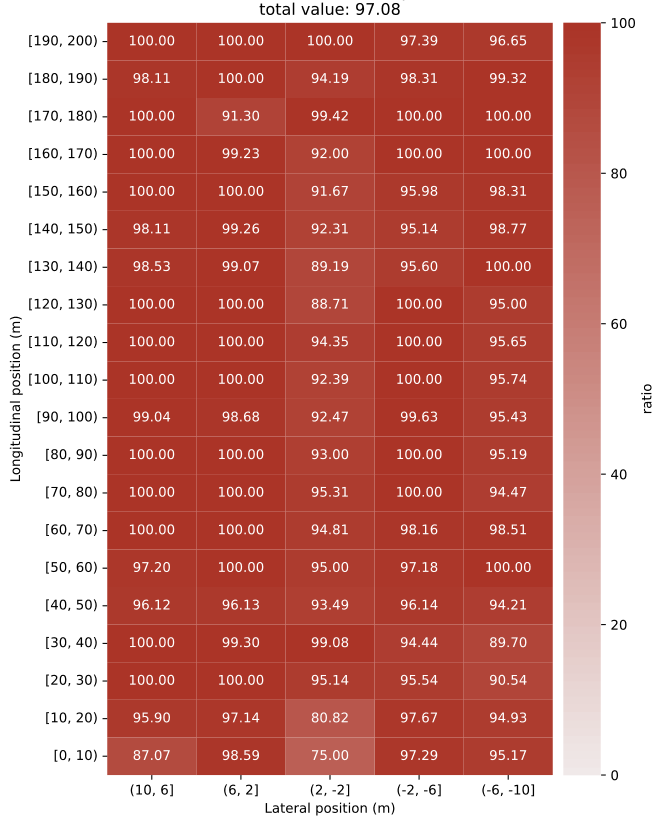}
  \end{subfigure}
  \hfill
  \begin{subfigure}{0.49\columnwidth}
  \includegraphics[width=\linewidth]{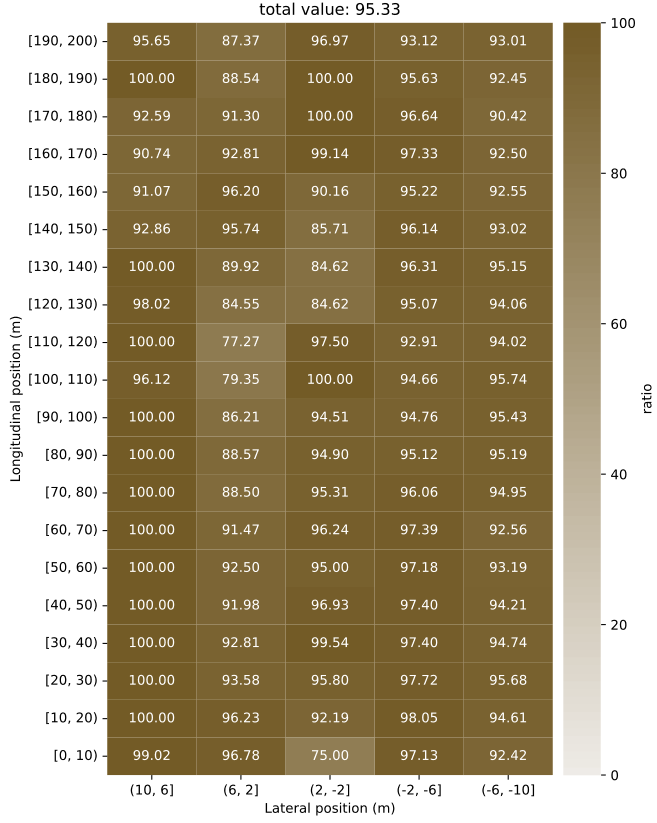}
  \end{subfigure}
  \caption{Precision and recall (in 4 m x 10 m blocks until 200 meters) of the proposed algorithm measured on our manually annotated \textbf{in-house} traffic sign dataset. \textbf{Left}: Precision (\textbf{97.08 \%}). \textbf{Right}: Recall (\textbf{95.33 \%}).}
  \label{fig:ts_metrics_pr}
\end{figure}

We also evaluated the localization errors of true positive detections using the absolute mean distance between the 3D bounding box centers and the annotations. Moreover, the absolute orientation error of the annotations is also evaluated. Our algorithm achieves low localization (\textbf{0.3 $\pm$ 0.16 meters}) and orientation (\textbf{11.09 degrees}) errors that are similar to the values measured on the Waymo dataset. Detailed metrics are shown in Fig. \ref{fig:ts_metrics} and Fig. \ref{fig:ts_metrics_pr}.

\begin{table}[t]
\begin{center}
\caption{Quantitative evaluation results of our automatic annotation method for traffic signs on in-house dataset.}
\label{tab:ts_validation_results}
\begin{tabular}{ll}
\hline\noalign{\smallskip}
Metric & Result\\
\noalign{\smallskip}
\hline
\noalign{\smallskip}
Association precision  & 97.08 \%\\
Association recall  & 95.33 \%\\
Localization error  & 0.30 $\pm$ 0.16 meters\\
Orientation error  &  11.09 degrees\\
\hline
\end{tabular}
\end{center}
\end{table}

\subsection{Validation of Automatic Traffic Light Annotation on in-house dataset}

\begin{figure}
\centering
\includegraphics[height=4.5cm]{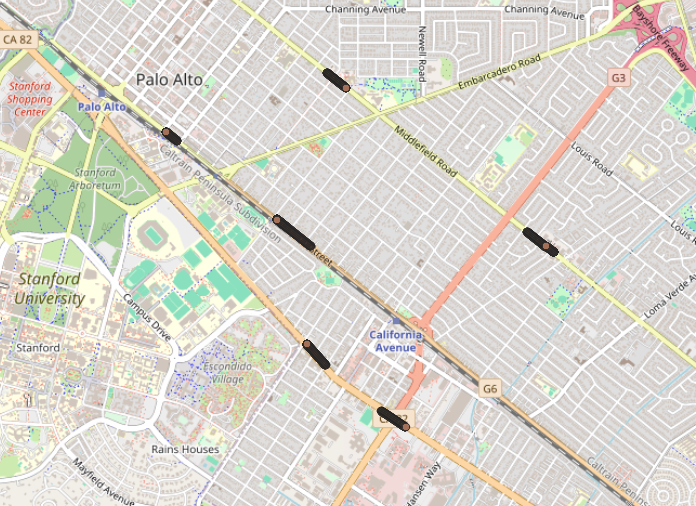}
\caption{Visualization of the traffic light validation route.}
\label{fig:tl_validation_route}
\end{figure}

\begin{figure}
    \centering
  \begin{subfigure}{0.49\columnwidth}
  \includegraphics[width=\linewidth]{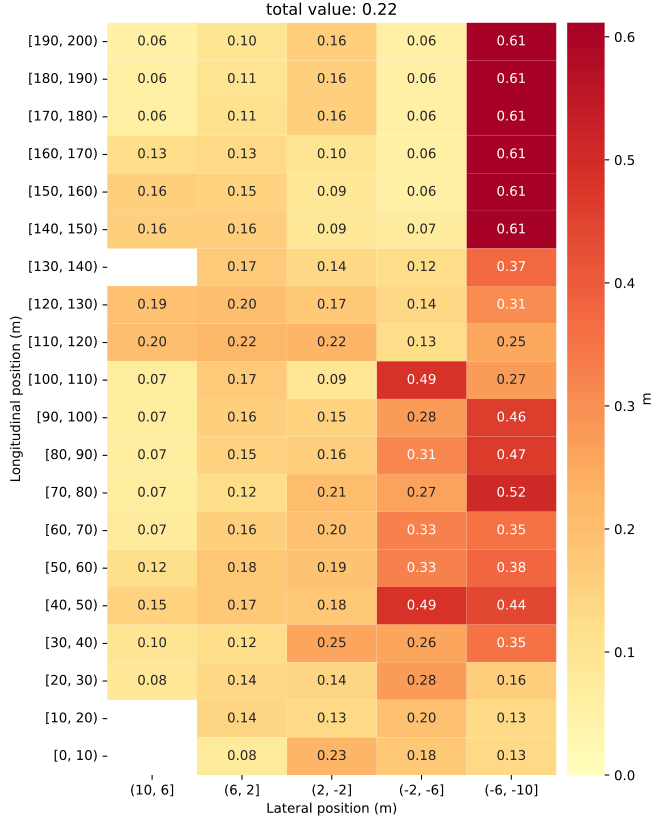}
  \end{subfigure}
  \hfill
  \begin{subfigure}{0.49\columnwidth}
  \includegraphics[width=\linewidth]{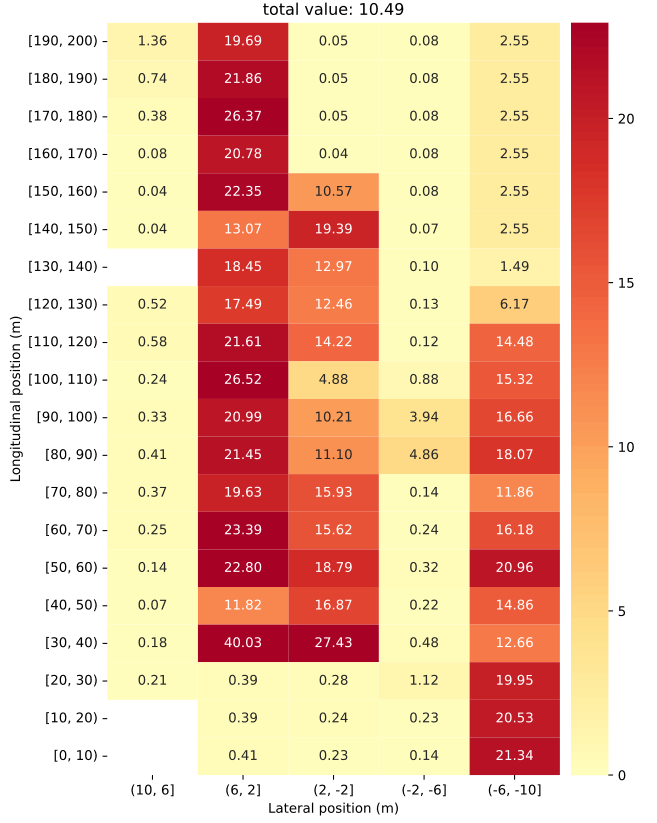}
  \end{subfigure}
  \caption{Evaluation results (in 4 m x 10 m blocks until 200 meters) of the proposed algorithm measured on our manually annotated \textbf{in-house} traffic light dataset. \textbf{Left}: Mean error in bounding box center estimation (\textbf{0.22 meters}). \textbf{Right}: Mean absolute error in box orientation (\textbf{10.49 degrees}).}
  \label{fig:tl_metrics}
\end{figure}

\begin{figure}
    \centering
  \begin{subfigure}{0.49\columnwidth}
  \includegraphics[width=\linewidth]{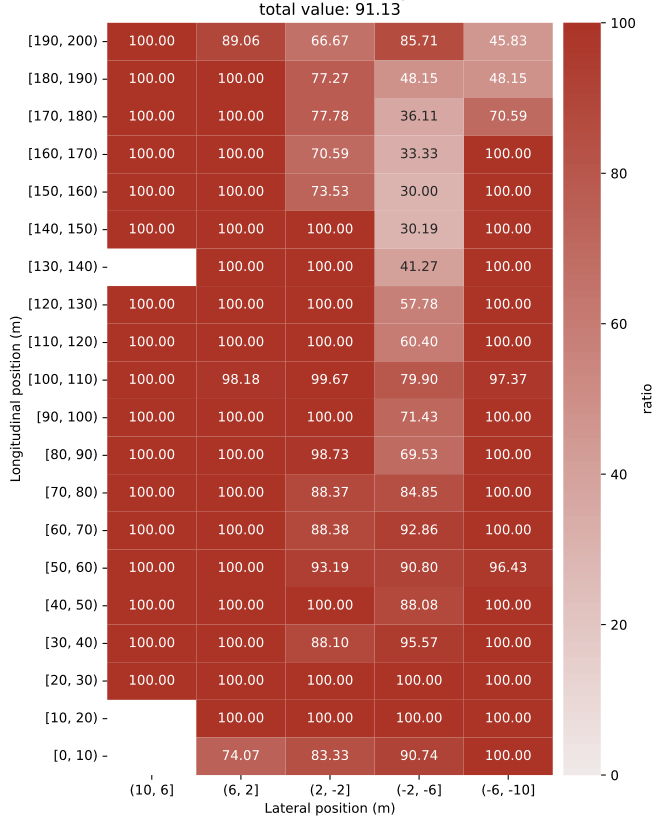}
  \end{subfigure}
  \hfill
  \begin{subfigure}{0.49\columnwidth}
  \includegraphics[width=\linewidth]{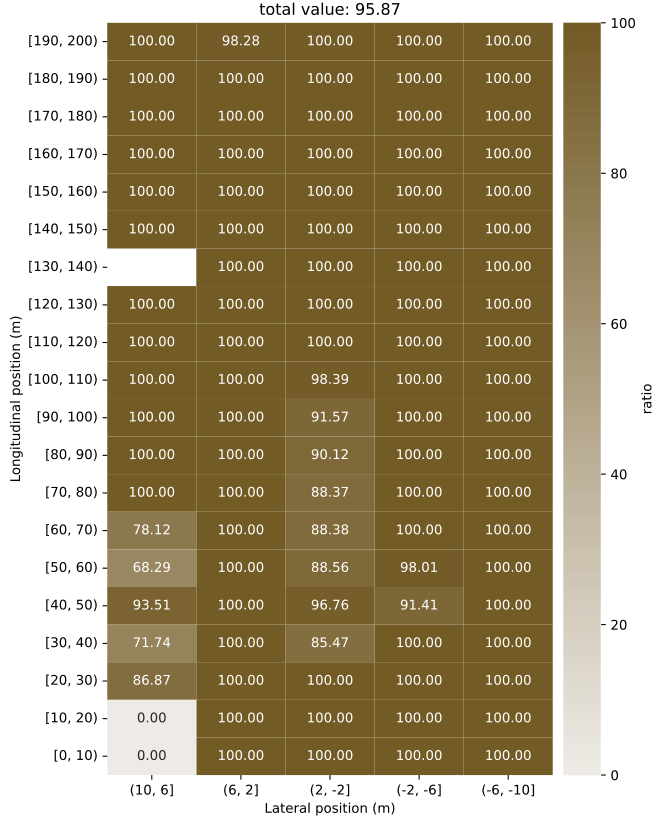}
  \end{subfigure}
  \caption{Precision and recall (in 4 m x 10 m blocks until 200 meters) of the proposed algorithm measured on our manually annotated \textbf{in-house} traffic light dataset. \textbf{Left}: Precision (\textbf{91.13 \%}). \textbf{Right}: Recall (\textbf{95.87 \%}).}
  \label{fig:tl_metrics_pr}
\end{figure}

We validated the automatic traffic light annotation algorithm at several intersections in Palo Alto, California. The validation route is approximately 1.3 kilometers long and includes 40 traffic lights (see the validation route in Figure \ref{fig:tl_validation_route}). The 3D bounding boxes of the traffic lights, as well as their states, were manually annotated. Consequently, we measured both localization performance and traffic light state classification accuracy. In the association metrics, a true positive means the prediction is within 1 meter of the ground truth and the predicted class is correct. All metrics were calculated within the range of [-10m, 10m] lateral and [0m, 200m] longitudinal positions of the instantaneous coordinate system. Our method achieved \textbf{91.13\%} precision and  \textbf{95.87\%} recall. The absolute localization error between the bounding box centers is \textbf{22 centimeters}, and the orientation absolute error is \textbf{10.49 degrees}. The traffic light color state classification accuracy is \textbf{94\%}. Detailed metrics are shown in Fig. \ref{fig:tl_metrics} and Fig. \ref{fig:tl_metrics_pr}.

\begin{table}[t]
\begin{center}
\caption{Quantitative evaluation results of our automatic annotation method for traffic lights on in-house dataset.}
\label{tab:tl_validation_results}
\begin{tabular}{ll}
\hline\noalign{\smallskip}
Metric & Result\\
\noalign{\smallskip}
\hline
\noalign{\smallskip}
Association precision  & 91.13 \%\\
Association recall  & 95.87 \%\\
Localization error  & 0.22 $\pm$ 0.20 meters\\
Orientation error  &  10.49 degrees\\
Color state classification accuracy & 94 \%\\
\hline
\end{tabular}
\end{center}
\end{table}

\section{Conclusion}

Despite self-driving developments that have been conducted for several decades, there is still no publicly available large-scale dataset with 3D annotated traffic lights and traffic signs. This indicates that annotating traffic management objects is challenging, even with manual resources. This is especially true for traffic lights, which are difficult to detect in LiDAR point clouds even for humans, as their physical characteristics (e.g., small size, high placement, and black coating) make it challenging for the sensor to produce easily detectable reflections. In this work, we developed a fully automated method to generate temporally consistent 3D bounding boxes with high localization precision for traffic lights and traffic signs, which can be used to train image-based perception models for self-driving cars. Additionally, we released a public dataset generated by our algorithm, available under a CC BY-NC-SA 4.0 license, allowing the research community to use it for non-commercial research purposes\footnote{\href{https://github.com/aimotive/aimotive_tl_ts_dataset}{https://github.com/aimotive/aimotive\_tl\_ts\_dataset}}.

\textbf{Limitations} The dataset is automatically annotated and, despite our extensive quality assurance process aimed at minimizing errors, it is still subject to annotation errors. Furthermore, the validation dataset size is limited which might hinder to measure the generalization ability of the proposed method.

\textbf{Future work} In the future, we aim to increase the manually annotated validation set's size continually. Furthermore, the traffic light detection precision shall be investigated on a larger sample. 

{\small
\bibliographystyle{ieee_fullname}
\bibliography{egbib}
}

\end{document}